\documentclass[letterpaper, 10 pt, journal, twoside]{IEEEtran}
\usepackage{graphics} % for pdf, bitmapped graphics files
\usepackage{epsfig} % for postscript graphics files
\usepackage{times} % assumes new font selection scheme installed
\usepackage{amsmath} % assumes amsmath package installed
\usepackage{amssymb}  % assumes amsmath package installed
\usepackage{afterpage}

\usepackage{graphicx}
\usepackage{upgreek}
\usepackage{bbm}
\usepackage{dsfont}
\usepackage{caption}
\usepackage{subcaption}
\usepackage{color}

\usepackage{booktabs}
\usepackage{makecell}
\usepackage{titlesec}
\usepackage{wrapfig}
\usepackage{array}
\usepackage{multirow}
\usepackage{colortbl}
\usepackage{tabularx}
\usepackage{cite}

\usepackage{xcolor}
\usepackage{pifont}
\usepackage{hyperref}

\ifCLASSINFOpdf
\else
\fi
\begin{document}

\makeatletter
\let\@oldmaketitle\@maketitle% Store \@maketitle
\renewcommand{\@maketitle}{\@oldmaketitle % Update
  \centering
  \setcounter{figure}{0}% 
  \includegraphics[width=0.85\linewidth]{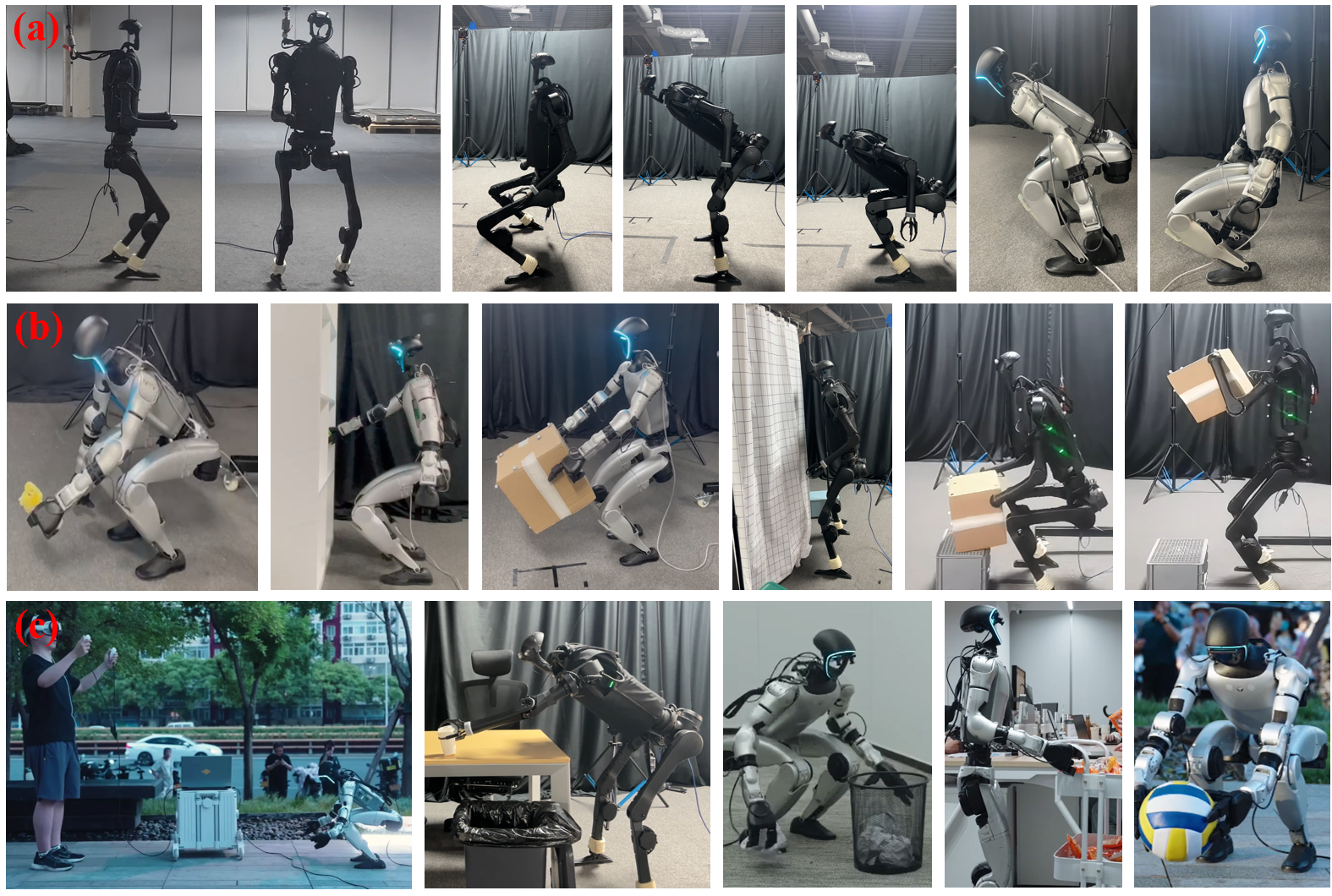}
  \captionof{figure}{\textbf{(a)} The humanoid showcases multiple real-world-ready primitive skills, including locomotion and body-pose-adjustment. \textbf{(b)} The humanoid autonomously accomplishes various WBC tasks. \textbf{(c)} The humanoid performs various tasks under our proposed teleoperation system.
  }
  \label{fig:teaser}
}
\makeatother

%
% paper title
% Titles are generally capitalized except for words such as a, an, and, as,
% at, but, by, for, in, nor, of, on, or, the, to and up, which are usually
% not capitalized unless they are the first or last word of the title.
% Linebreaks \\ can be used within to get better formatting as desired.
% Do not put math or special symbols in the title.
\title{Unleashing Humanoid Reaching Potential via Real-world-Ready Skill Space}
%
%
% author names and IEEE memberships
% note positions of commas and nonbreaking spaces ( ~ ) LaTeX will not break
% a structure at a ~ so this keeps an author's name from being broken across
% two lines.
% use \thanks{} to gain access to the first footnote area
% a separate \thanks must be used for each paragraph as LaTeX2e's \thanks
% was not built to handle multiple paragraphs
%

% \author{Michael~Shell,~\IEEEmembership{Member,~IEEE,}
%         John~Doe,~\IEEEmembership{Fellow,~OSA,}
%         and~Jane~Doe,~\IEEEmembership{Life~Fellow,~IEEE}% <-this % stops a space
% \thanks{M. Shell was with the Department
% of Electrical and Computer Engineering, Georgia Institute of Technology, Atlanta,
% GA, 30332 USA e-mail: (see http://www.michaelshell.org/contact.html).}% <-this % stops a space
% \thanks{J. Doe and J. Doe are with Anonymous University.}% <-this % stops a space
% \thanks{Manuscript received April 19, 2005; revised August 26, 2015.}}
\author{Zhikai Zhang$^{1,3*}$, Chao Chen$^{3,6*}$, Han Xue$^{1,3*}$, Jilong Wang$^{2,3}$, Sikai Liang$^{3,7}$, Yun Liu$^{1,3}$, Zongzhang Zhang$^{6}$, He Wang$^{2,3}$, and Li Yi$^{1,4,5}$%
\thanks{Manuscript received: July, 19, 2025; Revised October, 23, 2025; Accepted November, 25, 2025.}%Use only for final RAL version
\thanks{This paper was recommended for publication by Editor Abderrahmane Kheddar upon evaluation of the Associate Editor and Reviewers' comments.}
\thanks{*Zhikai Zhang, Chao Chen, and Han Xue are co-first authors.}
\thanks{$^{1}$First Author, Third Author, Sixth Author, and Ninth Author are with IIIS, Tsinghua University, China.}%
\thanks{$^{2}$Fourth Author and Eighth Author are with Peking University, China.}%
\thanks{$^{3}$First Author, Second Author, Third Author, Fourth Author, Fifth Author, Sixth Author, and Eighth Author are with Galbot, China.}%
\thanks{$^{4}$Ninth Author is with Shanghai AI Laboratory, China.}%
\thanks{$^{5}$Ninth Author is with Shanghai Qi Zhi Institute, China.}%
\thanks{$^{6}$Seventh Author is with Nanjing University, China.}%
\thanks{$^{7}$Fifth Author is with Tongji University, China.}%
\thanks{Digital Object Identifier (DOI): see top of this page.}
}
% note the % following the last \IEEEmembership and also \thanks - 
% these prevent an unwanted space from occurring between the last author name
% and the end of the author line. i.e., if you had this:
% 
% \author{....lastname \thanks{...} \thanks{...} }
%                     ^------------^------------^----Do not want these spaces!
%
% a space would be appended to the last name and could cause every name on that
% line to be shifted left slightly. This is one of those "LaTeX things". For
% instance, "\textbf{A} \textbf{B}" will typeset as "A B" not "AB". To get
% "AB" then you have to do: "\textbf{A}\textbf{B}"
% \thanks is no different in this regard, so shield the last } of each \thanks
% that ends a line with a % and do not let a space in before the next \thanks.
% Spaces after \IEEEmembership other than the last one are OK (and needed) as
% you are supposed to have spaces between the names. For what it is worth,
% this is a minor point as most people would not even notice if the said evil
% space somehow managed to creep in.

% The paper headers
%\markboth{Journal of \LaTeX\ Class Files,~Vol.~14, No.~8, August~2015}%
%{Shell \MakeLowercase{\textit{et al.}}: Bare Demo of IEEEtran.cls for IEEE Journals}
\markboth{IEEE Robotics and Automation Letters. Preprint Version. Accepted November, 2025}
{Zhang \MakeLowercase{\textit{et al.}}: Unleashing Humanoid Reaching Potential via Real-world-Ready Skill Space} 

% The only time the second header will appear is for the odd numbered pages
% after the title page when using the twoside option.
% 
% *** Note that you probably will NOT want to include the author's ***
% *** name in the headers of peer review papers.                   ***
% You can use \ifCLASSOPTIONpeerreview for conditional compilation here if
% you desire.

% If you want to put a publisher's ID mark on the page you can do it like
% this:
%\IEEEpubid{0000--0000/00\$00.00~\copyright~2015 IEEE}
% Remember, if you use this you must call \IEEEpubidadjcol in the second
% column for its text to clear the IEEEpubid mark.

% use for special paper notices
%\IEEEspecialpapernotice{(Invited Paper)}

% make the title area
\maketitle

% As a general rule, do not put math, special symbols or citations
% in the abstract or keywords.
\begin{abstract}
Humans possess a large reachable space in the 3D world, enabling interactions with objects at varying heights and distances.  However, realizing such large-space reaching on humanoids is a complex whole-body control (WBC) problem.
Learning from scratch often leads to optimization difficulty and poor sim2real transferability.
To address these challenges, we present Real-world-Ready Skill Space ($\text{R}^{2}\text{S}^{2}$), a structural skill prior that helps autonomous whole-body-control task execution in an efficient manner while maintaining sim2real transferability.
Inheriting knowledge from a set of real-world-ready primitive skills to ease multi-skill learning, $\text{R}^{2}\text{S}^{2}$ 
further expands the capability of primitive skills and learns a unified structural skill representation.
By sampling from $\text{R}^{2}\text{S}^{2}$, we unleash humanoid reaching potential in many real-world tasks. As a beneficial side effect, $\text{R}^{2}\text{S}^{2}$ can also support humanoid whole-body teleoperation with a large reachable space.
We validate the generalizability of $\text{R}^{2}\text{S}^{2}$ in various challenging goal-reaching tasks across different robot platforms, simulation and real world. We show some examples in Figure~\ref{fig:teaser}. Project page: \href{https://zzk273.github.io/R2S2/}{https://zzk273.github.io/R2S2/}.
\end{abstract}

% Note that keywords are not normally used for peerreview papers.
% \begin{IEEEkeywords}
% IEEE, IEEEtran, journal, \LaTeX, paper, template.
% \end{IEEEkeywords}

% For peer review papers, you can put extra information on the cover
% page as needed:
% \ifCLASSOPTIONpeerreview
% \begin{center} \bfseries EDICS Category: 3-BBND \end{center}
% \fi
%
% For peerreview papers, this IEEEtran command inserts a page break and
% creates the second title. It will be ignored for other modes.
\IEEEpeerreviewmaketitle

% The very first letter is a 2 line initial drop letter followed
% by the rest of the first word in caps.
% 
% form to use if the first word consists of a single letter:
% \IEEEPARstart{A}{demo} file is ....
% 
% form to use if you need the single drop letter followed by
% normal text (unknown if ever used by the IEEE):
% \IEEEPARstart{A}{}demo file is ....
% 
% Some journals put the first two words in caps:
% \IEEEPARstart{T}{his demo} file is ....
% 
% Here we have the typical use of a "T" for an initial drop letter
% and "HIS" in caps to complete the first word.
% \IEEEPARstart{T}{his} demo file is intended to serve as a ``starter file''
% for IEEE journal papers produced under \LaTeX\ using
% IEEEtran.cls version 1.8b and later.
% You must have at least 2 lines in the paragraph with the drop letter
% (should never be an issue)
\section{INTRODUCTION}

\IEEEPARstart{M}{any} human daily tasks can be viewed as reaching a series of points under certain conditions. A large reachable space enables interactions with objects at varying heights and distances—from overhead shelves to floor-level items. For humanoid robots to effectively assist humans in daily tasks, they should achieve a similar workspace~\cite{gu2025humanoid}. However, this presents a complex whole-body control challenge and requires mastering and intelligently utilize diverse skills—including base positioning and reorientation, height and body posture adjustments, and end-effector pose control within a dynamically unstable system. 

Traditional model-based control methods~\cite{lin2019efficient, ma2022combining} struggle with the inherent imperfections in system modeling and environmental disturbances. Recent end-to-end reinforcement learning has achieved great progress in humanoid whole-body control tasks~\cite{zhang2025track, he2024omnih2o, he2024learning, he2025asapaligningsimulationrealworld, xue2025unified}. Can we utilize it to endow humanoids the capability to accomplish tasks requiring a human-level large reachable space? 
We found that optimization and sim2real difficulty for such a complex whole-body control (WBC) problem are major concerns.
As mentioned before, multiple skills are required to be mastered and intelligently utilized for unleashing humanoid reaching potential. Learning all skills together from scratch is difficult. Existing works (e.g., AMO~\cite{li2025amoadaptivemotionoptimization} and HOMIE~\cite{ben2025homie}) often rely on highly intricate reward engineering and curriculum design to balance the rewards~\cite{ben2025homie} or trajectory optimization to provide guidance~\cite{li2025amoadaptivemotionoptimization}.
Additionally, a stable real-world performance usually requires iterative sim2real deployment to diagnose the sim2real gap and design corresponding constraints to mitigate the behavior discrepancy. For an end-to-end WBC policy, the coupling between different skills as well as between planning and control makes both sim2real diagnosis and constraint design much more challenging. Can we design prior knowledge to assist WBC tasks for less optimization and sim2real difficulty?

Toward this end, we propose \textbf{Real-world-Ready Skill Space ($\text{R}^{2}\text{S}^{2}$)}, aiming at constructing a skill space that encompasses and encodes various real-world-ready motor skills. Sampling from it, \textit{the learned space can serve as a structural skill prior and helps autonomous WBC task execution with minimal reward engineering efforts in a sim2real transferable manner.} 

To be specific, we first construct a library of pre-trained primitive skills to ease the optimization and sim2real transfer of multi-skill learning. These skills are task-agnostic and generalizable across different scenarios. 
Each skill can be individually tuned and sim2real evaluated for optimal real-world performance with minimal engineering efforts (because of decoupled training).
Though separated primitive skills can provide real-world-ready prior, they are insufficient to serve as a practical skill space for two reasons: 
1) separated training makes the coordination and transition between different skills out-of-distribution; 
2) different skills often have mismatched command spaces, lacking a unified representation for multi-skill planning. 

Therefore, we introduce a core stage called \textbf{heterogeneous skill ensembling}. At this stage, we inherit knowledge from pre-trained skills and expand it into a unified neural skill representation.
We achieve this by first constructing a heterogeneous skill training environment, then dynamically combining Imitation Learning (IL) and Reinforcement Learning (RL) to train a CVAE-based student policy, which inherits real-world-ready skill prior from the pre-trained teacher policies and explores new coordination and transition skills. 
Different from existing hierarchical humanoid control frameworks~\cite{li2025amoadaptivemotionoptimization, ben2025homie} where the planning is often conducted in the primary command space of the MLP-based low-level controller, the student network in our framework is designed as a CVAE to neurally model the motor skill distribution conditioned on proprioception, which proves to be a more efficient representation for multi-skill planning. 

With task-specific planning policies trained to sample from $\text{R}^{2}\text{S}^{2}$, our method enables the robot to autonomously accomplish various complex whole-body-control tasks in a sim2real transferable manner with minimal reward engineering efforts. In this work, we mainly focus on solving WBC tasks requiring a large reachable space. As a beneficial side effect, $\text{R}^{2}\text{S}^{2}$ also enables us to build a humanoid whole-body teleoperation system capable of reaching its full workspace, not just for table-top pick-and-place tasks.

We validate the generalizability of $\text{R}^{2}\text{S}^{2}$ across high-dof Unitree G1 (29 dofs) and full-sized Unitree H1 (1.8 meters tall) on various autonomous goal-reaching tasks. Extensive experiments are conducted in both simulation and real world to evaluate the effectiveness of our major designs. 

In summary, our main contributions are fourfold:
\begin{itemize}
    \item We propose \textbf{$\text{R}^{2}\text{S}^{2}$}, a structural skill prior that helps complex WBC task execution with minimal reward engineering efforts in a sim2real transferable manner.
    \item We propose a framework to construct $\text{R}^{2}\text{S}^{2}$. As the core, we propose \textbf{heterogeneous skill ensembling}, which can inherit knowledge from pre-trained skills and expand it into a unified skill representation.
    \item We implement $\text{R}^{2}\text{S}^{2}$ to unleash humanoid reaching potential in real world. As a beneficial side effect, we also utilize $\text{R}^{2}\text{S}^{2}$ to build a humanoid whole-body teleoperation system with a large reachable space. 
    \item We validate the generalizability of $\text{R}^{2}\text{S}^{2}$ across different humanoid platforms, tasks, sim and real.
    
\end{itemize}

\section{Related Works}
\label{sec:related}

% \begin{table}[htbp]
% \centering
% \vspace{3mm}
% \resizebox{0.45\textwidth}{!}{
%     \begin{tabular}{lllll}
% \hline
% L1      & L2 & Accuracy &  &  \\ \hline
% 0       &    &          &  &  \\
% Exbody2 &    &          &  &  \\
%         &    &          &  &  \\ \hline
% \end{tabular}
% \vspace{5mm}
% \caption{We evaluate the robustness of different $\lambda$ schedules.}
% \label{tab:lambda}
% \end{table}

\begin{table}[htbp]
\centering
\vspace{-3mm}
\begin{tabular}{cccc}
\hline
Methods  & Real-world Deploy    & Skill Prior & Skill Space \\ \hline
PULSE~\cite{luo2023universal}      & /   & Human Motion  &  Latent for RL    \\
HOMIE~\cite{ben2025homie} & Unitree G1   & Reward   &  Primary for IL   \\
AMO~\cite{li2025amoadaptivemotionoptimization}    &  Unitree G1  &  Traj. Opt. &  Primary for IL \\
Ours        &   Unitree G1 \& H1  & Primitive Skills &  Latent for RL  \\ \hline
\end{tabular}
\caption{Comparison with existing methods.}
\vspace{-3mm}
\label{tab:review}
\end{table}

\textbf{Humanoid Robot Learning.}
Reinforcement learning (RL) has achieved great progress in recent humanoid robot learning. Researches on locomotion~\cite{gu2024advancing,li2021reinforcement, duan2024learning, li2024reinforcement,wang2025beamdojo,xue2025unified} aim at provide bipedal humanoids with the ability to traverse different terrains in a stable and agile manner. But these works often focus only on the lower body of humanoids and ignore their whole-body reaching and interaction abilities. Learning-based humanoid whole-body-control~\cite{cheng2024expressive,ji2024exbody2,zhang2024wococo,dao2024sim,zhang2025track,he2024omnih2o,he2024learning,he2024hover,liu2024mimicking, zhuang2024humanoid, ben2025homie, xue2025unified, he2025asapaligningsimulationrealworld, li2025amoadaptivemotionoptimization} recently demonstrate new capabilities and push the boundaries of humanoid robots. 
In recent works AMO~\cite{li2025amoadaptivemotionoptimization} and HOMIE~\cite{ben2025homie}, humanoid reaching potential is unleashed by combining the capability of locomotion and body posture adjustment. However, these works either rely on highly intricate reward engineering and curriculum design to balance the rewards~\cite{ben2025homie} or trajectory optimization to provide guidance~\cite{li2025amoadaptivemotionoptimization}, making it hard to extend to new skills. In this work, we design a novel framework to ensemble multiple pre-trained skills without affecting the acquisition of each individual skill. Such a decoupling has the potential to scale up efficiently to more complex scenarios. 
Another key distinction between our method and AMO~\cite{li2025amoadaptivemotionoptimization} and HOMIE~\cite{ben2025homie} is that the latter mainly operate in the primary command space (e.g., root velocity, height) to collect teleoperation data for imitation learning, whereas our method focuses on constructing a skill space that enables efficient reinforcement learning exploration and exploitation without any human demonstrations as shown in Table~\ref{tab:review}.

\textbf{Skill Space Learning.}
In physics-based character animation, skill spaces~\cite{peng2022ase, won2022physics, yao2022controlvae, zhang2024freemotion, luo2023universal, luo2024grasping} are often learned to reuse motion prior from motion capture datasets. Motion imitation~\cite{won2022physics, zhang2024freemotion, luo2023universal, luo2024grasping } or adversarial learning~\cite{peng2022ase, juravsky2022padl, tessler2023calm} is used to form a skill latent space, and then the sampled latent variable can be translated into actions through a decoder. 

Though skill spaces constructed from human motion priors have been extensively studied in character animation, two main factors hinder the direct application of these methods to humanoid robots: 1) \textbf{Cross-embodiment gap.} Retargetting human motions to humanoids often require tremendous manual efforts to avoid artifacts; 2) \textbf{Training difficulty and poor sim2real transferability.} Skill priors learned from human motions are difficult to constrain and regularize though more diverse and expressive. Some motions may be challenging to transfer to real world. In comparison, our method acquires skill priors from real-world-ready primitive skills whose sim2real transferability is guaranteed as shown in Table~\ref{tab:review}.

\section{Unleashing Humanoid Reaching Potential via Real-world-Ready Skill Space}
\begin{figure*}
    \centering
    \vspace{0.1cm}
    \includegraphics[width=0.85\textwidth]{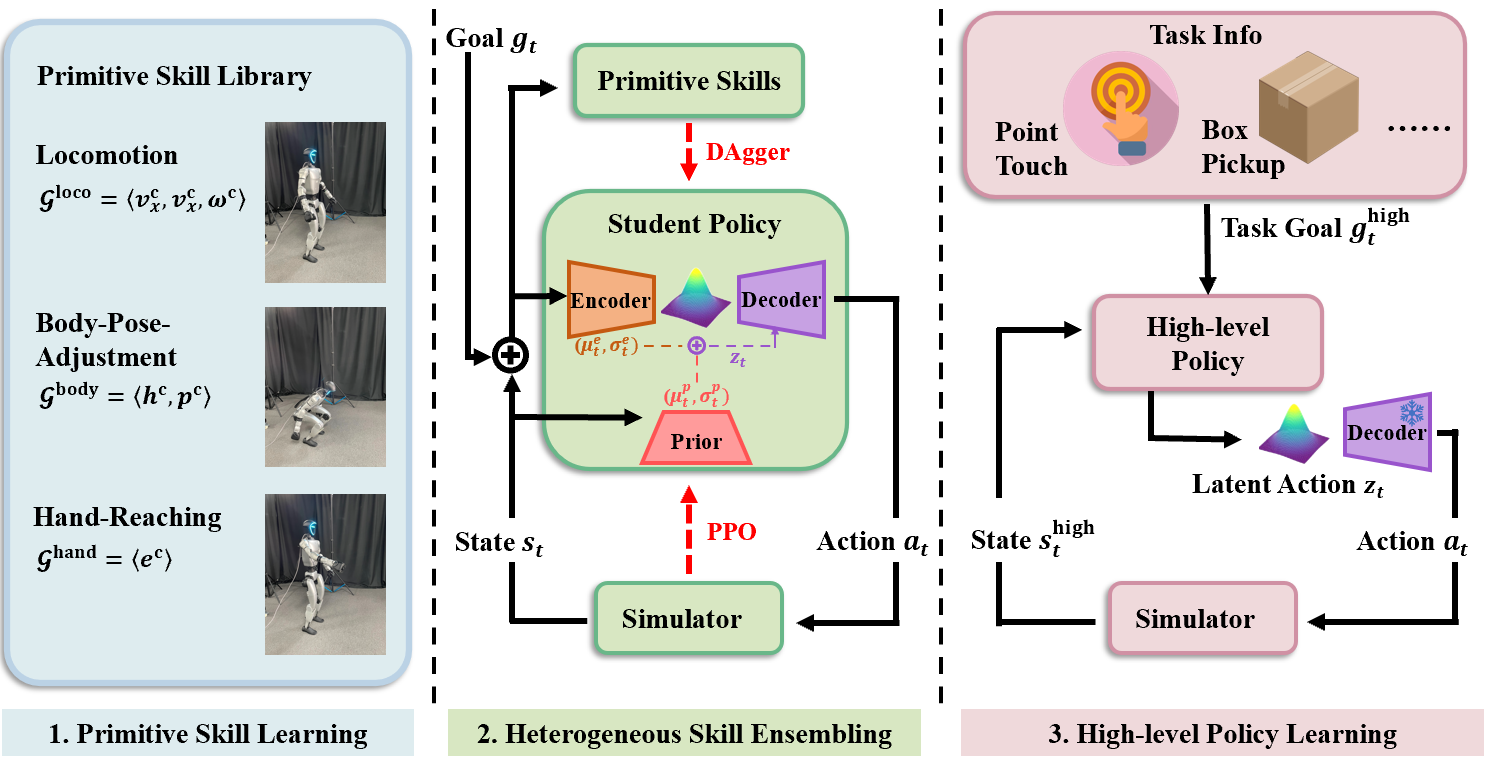}
    \caption{We present $\text{R}^{2}\text{S}^{2}$, a structural skill prior that helps autonomous WBC task execution in an efficient and sim2real transferable manner.}
    \label{fig:pipeline}
\vspace{-0.6cm}
\end{figure*}
In this section, we describe how to learn a \textbf{Real-world-Ready Skill Space ($\text{R}^{2}\text{S}^{2}$)} and utilize it to support practical WBC tasks, with a focus on unleashing humanoid reaching potential.
We first introduce the construction of primitive skill library in Section~\ref{sec:skill_library}. We then present the core \textbf{heterogeneous skill ensembling} stage in Section~\ref{sec:skill_space}. Finally, we show how to sample from $\text{R}^{2}\text{S}^{2}$ to efficiently solve various real-world goal-reaching tasks in Section~\ref{sec:planning}.
The pipeline is shown in Figure~\ref{fig:pipeline}.
In this work, we use PPO~\cite{ppo} for all of our policy training. We add random Gaussian noise to the policy observations, randomize dynamics parameters, and apply random external force disturbances for sim-to-real transfer. We use Isaac Gym~\cite{makoviychuk2021isaac} for simulation. We will now introduce the primitive skill library:

\begin{table}[t]
\centering
\renewcommand{\arraystretch}{1.2}
\setlength{\tabcolsep}{4pt}
\vspace{0.1cm}
\caption{We list all of our reward terms and corresponding weights here.}
\begin{tabular}{l l l}
\hline
\hline
\multicolumn{3}{c}{$r_{\text{command}}$} \\
\hline
\hline
\textbf{Term} & \textbf{Equation} & \textbf{Scale} \\
\hline
\textit{Locomotion} & & \\
\quad Linear velocity tracking & $\exp\{{-5.0 {|v^{c}-v|^2}\}}$ & 1.0 \\
\quad Angular velocity tracking & $\exp\{{-7.0{|\omega^{c}-\omega|^2}\}}$ & 1.0 \\
\textit{Body-Pose-Adjustment} & &  \\
\quad Body height tracking & $\exp\{{-4.0 {|h^{c}-h|^2}\}}$ & 1.0 \\
\quad Pitch angle tracking & $\exp\{{-4.0 {|b^{c}-b|^2}\}}$ & 1.0 \\
\textit{Hand-Reaching} & &  \\
\quad End-effector pose tracking & $\exp\{{-4.0 {|e^{c}-e|^2}\}}$ & 1.0 \\
\hline
\hline
\multicolumn{3}{c}{$r_{\text{behavior}}$} \\
\hline
\hline
\textbf{Term} & \textbf{Equation} & \textbf{Scale} \\
\hline
\textit{Locomotion} & & \\
\quad Gait velocity tracking &  $\sum_{\text{foot}}[1 - \text{C}_\text{foot}(t)] |v_\text{foot}|^2$ & 1.0 \\
\quad Gait force tracking &  $\sum_{\text{foot}}[\text{C}_\text{foot}(t)] |\text{f}_\text{foot}|^2$ & 1.0 \\
\textit{Body-Pose-Adjustment} & & \\
\quad Base roll error & $\exp\{{-4.0 {r^2}\}}$ & 1.0 \\
\quad Leg pos symmetry & $ \| q_{\text{left\_leg}} - q_{\text{right\_leg}} \|_2$ & 0.5 \\
\quad Leg torque symmetry & ${ {|a_{\text{low}}^{\text{left\_leg}}-a_{\text{low}}^{\text{right\_leg}}|}}$ & -0.2 \\
\quad Contact ground & $ c_{\text{left}} * c_{\text{right}} $ & 1.0 \\
\hline
\hline
\multicolumn{3}{c}{$r_{\text{regularization}}$} \\
\hline
\hline
\textbf{Term} & \textbf{Equation} & \textbf{Scale} \\
\hline
\quad Action acc & $ \| a_t - 2a_{t-1} + a_{t-2}\|_2$ & -0.01 \\
\quad Action rate & $ \| a_t - a_{t-1}\|_2$ & -0.01 \\
\quad Collision & undesired collision & -5.0 \\
\quad Default joint error &  $\exp\{{-2.0{|q-q_0|^2}\}}$ & 0.2 \\
\hline
\hline
\multicolumn{3}{c}{$r_{\text{task}}$} \\
\hline
\hline
\textbf{Term} & \textbf{Equation} & \textbf{Scale} \\
\hline
\textit{Single-point Touch} & & \\
\quad Single-point touch & $\exp\{{-{dist(hand, point)}\}}$ & 1.0 \\
\textit{Dual-point Touch} & & \\
\quad Dual-point touch & $\exp\{{-{dist(hands, points)}\}}$ & 1.0 \\
\textit{Shelf Touch} & & \\
\quad Shelf touch & $\exp\{{-{dist(hand, point)}\}}$ & 1.0 \\
\textit{Box Pickup} & & \\
\quad Hand approach & $\exp\{{-{dist(hand, box\_side)}\}}$ & 1.0 \\
\quad Lift box & $\exp\{{-{dist_{z}(box\_height, 1.4)}\}}$ & 1.0 \\
\hline
\end{tabular}
\label{table:rew}
\vspace{-5mm}
\end{table}

\subsection{Primitive Skill Library}
\label{sec:skill_library}
Aiming at unleashing humanoid reaching potential, we design the primitive skill library $\{\pi_{i}^{\text{prim}}\}_{i=1}^{n}$ as 
\textcolor{orange}{locomotion}, \textcolor{magenta}{body-pose-adjustment} (changing body height, bending over) and \textcolor{cyan}{hand-reaching}, 
which can handle with many goal-reaching scenarios. Compared with training a generalist controller from scratch, such a decomposition avoids performance degradation brought by multi-skill learning.

Our primitive skills can be seen as goal-conditioned RL policies \(\pi^{\text{prim}}: \mathcal G^{\text{prim}} \times \mathcal S^{\text{prim}} \mapsto \mathcal A^{\text{prim}}\), where \(\mathcal G^{\text{prim}}\) includes goal commands $g_t$ specifying skill target. 
\(\mathcal S^{\text{prim}}\) includes the robot's proprioceptive observation and history action information $s_t=[\omega_{t}, gr_t,  q_t, \dot{q}_t, a_{t-1}]$ at each timestep $t$, where $\omega_{t}, gr_t,  q_t, \dot{q}_t, a_{t-1}$ are angular velocity in the base frame, projected gravity, body-part dof positions, body-part dof velocities, and last-frame low-level action, respectively. It is worth noting that for $q_{t}, \dot{q}_t, a_{t-1} $, each skill policy only takes relevant body part information as observation, lower-body for locomotion and body-pose-adjustment and two arms for hand-reaching.
% ~\eric{provide an example for intuitive understanding}. 
\(\mathcal A^{\text{prim}}\) includes the robot body-part action (PD targets) $a^{\text{prim}}$, which is fed into a PD controller for torque computation. $a^{\text{prim}}$ only controls corresponding body part for each skill, and other joints are fixed. Their training reward can be written as:
\begin{equation}
    r_{\text{prim}} = r_{\text{command}} + r_{\text{behavior}} + r_{\text{regularization}}, 
\end{equation}
where $r_{\text{command}}$ represents skill command tracking objectives, $r_{\text{behavior}}$ 
depicts skill-specific behavior constraints for sim2real stability, and $r_{\text{regularization}}$ is skill-agnostic regularization. In the following sections, we mainly introduce $r_{\text{behavior}}$ of each skill since they are most important for sim2real transfer. For detailed rewards, please refer to Table~\ref{table:rew}.

\textbf{Locomotion.}
For locomotion, \(\mathcal G^{\text{loco}} = \langle v_x^{\text{c}}, v_y^{\text{c}}, \omega^{\text{c}} \rangle\) actuates the humanoid to track desired linear and angular velocities of the robot base in the robot base frame. To constrain locomotion behavior and replicate human-like bipedal gaits, we model each foot’s motion as an alternating sequence of swing and stance phases and introduce a periodic reward framework inspired by~\cite{siekmann2021sim, margolis2023walk}:
\begin{equation}
r^{\text{loco}}_{\text{behavior}} = r_{\text{gait\_velocity}} + r_{\text{gait\_force}},
\end{equation}
\begin{equation}
r_{\text{gait\_velocity}} = \sum_{\text{foot}}[1 - \text{C}_\text{foot}(t)] |v_\text{foot}|^2, 
\end{equation}
\begin{equation}
r_{\text{gait\_force}} = \sum_{\text{foot}}[\text{C}_\text{foot}(t)] |\text{f}_\text{foot}|^2,
\end{equation}
where $ C_{\text{foot}}(t) $ follows Von Mises distributions and $t \in [0, 1)$ is a time-dependent phase variable cycling periodically through normalized time.

\textbf{Body-Pose-Adjustment.}
For body-pose-adjustment, \(\mathcal G^{\text{body}} = \langle h^{\text{c}}, b^{\text{c}} \rangle\) tracks the base height and torso bending angle in the global frame. 
We found that for such a skill, kinematic and dynamic symmetry is important for real-world stability, so we introduce:
\begin{equation}\label{eq:behavior_body}
    r^{\text{body}}_{\text{behavior}} = r_{\text{base\_roll}} + r_{\text{leg\_pos}} + r_{\text{leg\_torque}} + r_{\text{touch\_ground}}, 
\end{equation}
where $r_{\text{base\_roll}}$ and $r_{\text{leg\_pos}}$ are designed for kinematic symmetry and $r_{\text{leg\_torque}}$ and $r_{\text{touch\_ground}}$ are designed for dynamic balance. For more details, please refer to Table~\ref{table:rew}.

\textbf{Hand-Reaching.} 
For hand reaching, \(\mathcal G^{\text{hand}} = \langle e^{\text{c}} \rangle\) tracks the target end-effector 6D pose in the robot local frame. Arms are relatively easy for sim2real deployment, so we do not specifically design any $r_{\text{behavior}}$ for this skill.

\subsection{Heterogeneous Skill Ensembling}
\label{sec:skill_space}
Given real-world-ready primitive skills $\{\pi_{i}^{\text{prim}}\}_{i=1}^{n}$, a straight attempt to reuse these primitive skills for WBC tasks is directly planning in their primary command spaces, for example, training a planner policy to predict which primitive skill to be activated at each timestep and output corresponding primary commands (e.g., $v_x^{\text{c}}, v_y^{\text{c}}, \omega^{\text{c}}$ for locomotion).
But individual primitive skills are actually insufficient for practical WBC tasks. Because of separated training, isolated primitive skills are unseen to each other. 
The coordination (e.g., upper-body reaching an object while lower-body squatting) and transition (e.g., lower-body from locomotion to body-pose-adjustment) between different skills are out-of-distribution problems. Naïvely concatenating actions of different body parts or switching from locomotion to body-pose-adjustment skill will lead to instability or even cause robot to fall. Without seamless coordination and transition, the skill space is incomplete for practical task accomplishment. 
In addition, the command space of primitive skills ($v_x^{\text{c}}, v_y^{\text{c}}, \omega^{\text{c}}$ for locomotion, $h^{\text{c}}, b^{\text{c}}$ for body-pose-adjustment, and $e^{\text{c}}$ for hand-reaching in our setting) are mismatched. Direct combination of skill indicators and corresponding skill commands is not a structural action space for planner policy and poses challenge for multi-skill planning. These two drawbacks make the primitive skills inefficient for planning as shown in~Section~\ref{sec:rrss}.

To solve these problems, we propose to train an ensemble student policy $\pi^{\text{ensem}}(a_{t}|s_{t},g_{t})$ with a variational information bottleneck to ensemble heterogeneous skills. ``Ensemble" means not only imitating different primitive skills, but also expanding their coordination and transition capability. During heterogeneous skill ensembling, different skills are encoded into a unified latent skill space $z$, and then decoded into per-joint actions.

\textbf{Expanding Coordination and Transition Capability.}
The ensemble student policy $\pi^{\text{ensem}}(a_{t}|s_{t},g_{t})$ should not only retain the real-world transferability of primitive skills, but also expand new coordination and transition skills.
To retain the real-world transferability of primitive skills, an effective choice is to leverage online imitation learning methods (e.g., DAgger~\cite{ross2011reduction}). In particular, we first construct a heterogeneous skill training environment to model skill coordination and transition: 1) we simultaneously send goal commands for different body parts (e.g., the policy needs to track target hand 6D pose while walking at the same time) to model skill coordination; 2) we randomly set the skill of a certain body part to transition from one to another in an episode to model skill transition. At each timestep $t$, two primitive skills $\{\pi_{t}^{\text{lower}}, \pi_{t}^{\text{upper}}\}, \pi_{t}^{\text{lower}} \in \{\pi^{\text{loco}}, \pi^{\text{body}}\}$ and $ \pi_{t}^{\text{upper}} \in \{ \pi^{\text{hand}}\}$, serve as teacher policies for different body parts, one for lower-body and the other one for upper-body. A skill indicator is included in student policy goal $g_{t}$ to indicate which teacher policy is activated. When transition happens, we let $\pi_{t+1}^{\text{lower}} \neq \pi_{t}^{\text{lower}}$.
By doing so, all possible coordination and transition situations are covered in the student policy training process.
 
However, relying only on imitation learning can not expand student policies with new capabilities (e.g., coordination and transition between different skills) beyond teacher policies. Thus, we propose to combine imitation learning and reinforcement learning by dynamically adding IL loss and RL loss together.
The IL, which is DAgger in our setting, distills real-world-ready skill prior from multiple teacher policies. Based on this, the RL, which is PPO in our setting, further encourages the policies to learn new behaviors for seamless transition and coordination. The reward function can be written as:
\begin{equation}
\mathcal{L_{\text{Ensem}}} = \lambda_{1} \mathcal{L}_{\text{DAgger}} + \lambda_{2} \mathcal{L}_{\text{PPO}},
\end{equation}
where $\lambda_{1}$ decreases from 0.95 to 0.05 gradually and $\lambda_{2}$ inversely adjusted. This design encourages the student policy to mimic teacher policies first and exploring new behaviors latter. Instead of utilizing a sequential training strategy that first employs IL for pretraining followed by exclusive RL fine-tuning, we maintain the supervision signal from the teacher policies throughout the entire training process. 
Our strategy can prevent catastrophic forgetting of the skill prior provided by teacher policies.
We let
\begin{equation}
\mathcal{L}_{\text{DAgger}} = \mathbb{E}_{(s, a^*) \sim \mathcal{D}_{\text{agg}}} \left[ \| a^{\text{ensem}}_{t} - a_{t}^* \|^2 \right], 
\end{equation}
where $a^{\text{ensem}}_{t}$ is the output action of $\pi^{\text{ensem}}$ and $a_{t}^* = concat(a^{\text{lower}}_{t}, a^{\text{upper}}_{t})$ is the combination of actions from lower-body and upper-body teacher policies. For $\mathcal{L}_{\text{PPO}}$, we simply combine the reward terms of $\pi^{\text{lower}}_{t}$ and $\pi^{\text{upper}}_{t}$ defined in the primitive skill training stage. We found that the student policy can successfully learn coordination and transition skills without any additional reward terms. Though coordination and transition are newly learned at this stage, the skill prior inherited from teacher policies serves as a good warm-up and makes the new capabilities sim2real transferable.

\textbf{Learning a Unified Neural Skill Representation.}
While the student policy can ensemble multiple primitive skills, mismatched command spaces hinders efficient high-level planning due to the absence of a unified skill representation. Concretely, if we train the high-level planner to output all the commands of different primitive skills simultaneously, it can easily cause conflicts (e.g., asking the robot to walk fast while crouching very low). Alternatively, if the high-level planner outputs a skill indicator along with the corresponding skill command, we find that the discrete skill indicator hinders RL exploration and leads the planner to consistently favor one particular skill.
To mitigate this, we adopt an encoder-decoder framework with a conditional variational information bottleneck to encode skills in a continuous and unified skill space inspired by~\cite{luo2023universal}, which includes a variational encoder $\mathcal{E}(z_t | s_t, g_t) = \mathcal{N}(z_t; \mu^e(s_t, g_t), \sigma^e(s_t, g_t))$ to model latent codes conditioned on current state and goal, a decoder $\mathcal{D}(a_t | s_t, z_t)$ maps the sampled latent code to action conditioned on state, and a learnable conditional prior $\mathcal{P}(z_t | s_t) = \mathcal{N}(z_t; \mu^p(s_t), \sigma^p(s_t))$ to capture state-based action distribution instead of assuming a fixed unimodal Gaussian.
The total loss in training $\pi^{\text{ensem}}$ can be written as:
\begin{equation}
\mathcal{L}_{\text{Total}} = \mathcal{L}_{\text{Ensem}} + \lambda_{3} \mathcal{L}_{\text{Regu}} + \lambda_{4} \mathcal{L}_{\text{KL}},
\end{equation}
where
\begin{equation}
\mathcal{L}_{\text{Regu}} = \| \mu^e(s_t, g_t) - \mu^e(s_{t+1}, g_{t+1}) \|
\end{equation}
encourages temporal consistency between consecutive latent codes and makes the skill space more structural. Since we allow primitive skills to transition from one to another (e.g., from $\pi^{\text{loco}}$ to $\pi^{\text{body}}$) during this ensembling stage, two different primitive skills and their transition can be modeled as a continuous distribution in the latent space with $\mathcal{L}_{\text{Regu}}$. $\mathcal{L}_{\text{KL}} = D_{\text{KL}}(\mathcal{E}(z_t | s_t, g_t) \parallel \mathcal{P}(z_t | s_t))$ encourages the distribution of the latent code to be close to the learnable prior. 

\subsection{High-Level Planning in Real-World-Ready Skill Space}
\label{sec:planning}
In this work, we mainly focus on utilizing $\text{R}^{2}\text{S}^{2}$ to solve WBC tasks requiring a large reachable space. The skill prior encoded in $\text{R}^{2}\text{S}^{2}$ helps autonomous task execution in a sim2real transferable manner with simple task requirement inputs.
We achieve this by training task-specific high-level planners $\pi^{\text{plan}}(z_t | s^{\text{plan}}_t, g^{\text{plan}}_t)$ with RL to sample from the learned latent skill space. 
The action for $\pi^{\text{plan}}$ is now in the latent $z_t$ space. The sampled $z_t$ is decoded into per-joint actions $a_t$ via the frozen decoder $\mathcal{D}$. The training reward can be written as:
\begin{equation}
    r_{\text{plan}} = r_{\text{task}} + r_{\text{regularization}}, 
\end{equation}
where $r_{\text{task}}$ is task execution objective describing desired task requirements.
$r_{\text{regularization}}$ is the skill-agnostic regularization reward reused from the skill library construction stage to enhance motion stability. It is worth noting that with $\text{R}^{2}\text{S}^{2}$, we only need to define $r_{\text{task}}$ at this stage. The humanoid can autonomously accomplish various tasks in a sim2real transferable manner without any additional designs. For detailed $r_{\text{task}}$, please refer to Table~\ref{table:rew}.
\section{Experiments}
In this section, comprehensive experiments in both simulation and real-world will be conducted to answer the following questions: \textbf{Q1.} (Section~\ref{sec:unleash}) Compared with baseline methods, can Real-world-Ready Skill Space ($\text{R}^{2}\text{S}^{2}$) better assist with various WBC tasks in an efficient and sim2real transferable manner? \textbf{Q2.} (Section~\ref{sec:rrss}) How does each part of $\text{R}^{2}\text{S}^{2}$ contribute to the final results? The quantitative results are reported on Unitree H1.

\begin{table*}
\centering
    \renewcommand{\arraystretch}{1.2}
    \vspace{0.2cm}
    \caption{We compare $\text{R}^{2}\text{S}^{2}$ with baseline methods. ``SR" is short for \textbf{Success Rate} and ``DE" is short for \textbf{Distance Error}. Our method exceeds baselines in task accomplishment performance and sim2real transferability.}
    \begin{tabular}
    {@{\extracolsep{\fill}} p{0.2\linewidth} 
    >{\centering\arraybackslash}p{0.06\linewidth} 
    >{\centering\arraybackslash}p{0.06\linewidth} 
    >{\centering\arraybackslash}p{0.06\linewidth}
    >{\centering\arraybackslash}p{0.06\linewidth} 
    >{\centering\arraybackslash}p{0.06\linewidth} 
    >{\centering\arraybackslash}p{0.06\linewidth} 
    >{\centering\arraybackslash}p{0.06\linewidth}
    >{\centering\arraybackslash}p{0.06\linewidth} 
    >{\centering\arraybackslash}p{0.1\linewidth}
    } 
    \hline
    \multirow{2}{*}{Method} & \multicolumn{2}{c}{Single-point Touch} & \multicolumn{2}{c}{Dual-point Touch} & \multicolumn{2}{c}{Shelf Touch} & \multicolumn{2}{c}{Box Pickup} & \multirow{2}{*}{Sim2Real}\\
    \cline{2-9} 
    & {\textit{SR(\%)} $\uparrow$} & {\textit{DE(m)} $\downarrow$} 
    & {\textit{SR(\%)} $\uparrow$} & {\textit{DE(m)} $\downarrow$} 
    & {\textit{SR(\%)} $\uparrow$} & {\textit{DE(m)} $\downarrow$} 
    & {\textit{SR(\%)} $\uparrow$} & {\textit{DE(m)} $\downarrow$}  \\ 
     \hline
    \textit{Sim} \\
    \quad Vanilla PPO~\cite{ppo} & 11.63 & 0.49 & 5.71 & 0.52 & 13.20 & 0.37 &0.00 &0.65 &\textcolor{red}{\ding{55}} \\ 
    \quad Vanilla DreamerV3~\cite{hafner2023mastering} & 30.51 & 0.15 & 22.83 & 0.33 & 8.93& 0.28& 0.00& 0.59&\textcolor{red}{\ding{55}} \\
    \quad HumanoidBench~\cite{sferrazza2024humanoidbench} & \textbf{100} & 0.04 & \textbf{100} & 0.04 & 47.69 & 0.13 & 0.00 & 0.44 &\textcolor{red}{\ding{55}} \\
    \quad \textbf{Ours} & \textbf{100} & \textbf{0.03} & \textbf{100} & \textbf{0.03} & \textbf{100} & \textbf{0.02} & \textbf{100} & \textbf{0.04} &\textcolor{green}{\ding{51}} \\ 
    \hline
    \textit{Real} \\
    \quad \textbf{Ours} & \textbf{100} & \textbf{0.03} & \textbf{100} & \textbf{0.03}& \textbf{100} & \textbf{0.03} & \textbf{90} & \textbf{0.08} &N/A\\
    \hline
    \end{tabular}
    \label{table:unleash}
\end{table*}

\begin{table*}
\centering
    \renewcommand{\arraystretch}{1.2}
    \caption{We evaluate the effectiveness of our major designs. ``SR" is short for \textbf{Success Rate} and ``DE" is short for \textbf{Distance Error}. Each of our major designs contributes to the final results.}
    \begin{tabular}
    {@{\extracolsep{\fill}} p{0.2\linewidth} 
    >{\centering\arraybackslash}p{0.06\linewidth} 
    >{\centering\arraybackslash}p{0.06\linewidth} 
    >{\centering\arraybackslash}p{0.06\linewidth}
    >{\centering\arraybackslash}p{0.06\linewidth} 
    >{\centering\arraybackslash}p{0.06\linewidth} 
    >{\centering\arraybackslash}p{0.06\linewidth} 
    >{\centering\arraybackslash}p{0.06\linewidth}
    >{\centering\arraybackslash}p{0.06\linewidth} 
    >{\centering\arraybackslash}p{0.1\linewidth}
    } 
    \hline
    \multirow{2}{*}{Method} & \multicolumn{2}{c}{Single-point Touch} & \multicolumn{2}{c}{Dual-point Touch} & \multicolumn{2}{c}{Shelf Touch} & \multicolumn{2}{c}{Box Pickup} & \multirow{2}{*}{Sim2Real}\\
    \cline{2-9} 
    & {\textit{SR(\%)} $\uparrow$} & {\textit{DE(m)} $\downarrow$} 
    & {\textit{SR(\%)} $\uparrow$} & {\textit{DE(m)} $\downarrow$} 
    & {\textit{SR(\%)} $\uparrow$} & {\textit{DE(m)} $\downarrow$} 
    & {\textit{SR(\%)} $\uparrow$} & {\textit{DE(m)} $\downarrow$}  \\ 
     \hline
    $\text{R}^{2}\text{S}^{2}$ w/o PS & 49.58 & 0.14 & 43.15 & 0.12 & 47.81 & 0.22 &30.74 & 0.26&\textcolor{red}{\ding{55}} \\
    $\text{R}^{2}\text{S}^{2}$ w/o SE & 30.51 & 0.15 & 26.39 & 0.16 & 28.34& 0.29& 22.82& 0.33&\textcolor{red}{\ding{55}} \\
    $\text{R}^{2}\text{S}^{2}$ w/ Seq-ILRL & 47.16 & 0.10 & 45.98 & 0.09 & 54.72 & 0.09 & 29.53 & 0.24 &\textcolor{red}{\ding{55}} \\
    $\text{R}^{2}\text{S}^{2}$ w/o LS & 56.87 & 0.10 & 52.38& 0.07 & 44.86 & 0.17 & 43.29 & 0.19 &\textcolor{green}{\ding{51}} \\
    \textbf{Ours} & \textbf{100} & \textbf{0.03} & \textbf{100} & \textbf{0.03} & \textbf{100} & \textbf{0.02} & \textbf{100} & \textbf{0.04} &\textcolor{green}{\ding{51}} \\ 
    \hline
    \end{tabular}
    \label{table:rrss}
\vspace{-5mm}
\end{table*}

\subsection{Performance of $\text{R}^{2}\text{S}^{2}$ on Whole-Body Control Tasks}
\label{sec:unleash}
In this part, we want to compare our method with baseline methods to evaluate whether $\text{R}^{2}\text{S}^{2}$ assists with humanoid whole-body control tasks. We select four WBC tasks that require unleashing humanoid reaching potential and compare the performance of different methods on them.
\subsubsection{Experiment Setting}
In our experiment setting, we select four representative goal-reaching tasks, including scenarios involving a single hand, both hands, obstacle avoidance, and humanoid-object interaction: 
\begin{itemize}
    \item \texttt{Single-point Touch:} We randomly set one point within a $2m \times 2m$ square in front of the robot, with a height ranging from 0.1 meters to 2.0 meters. The humanoid is asked to touch the point with one hand. 
    \item \texttt{Dual-point Touch:} We randomly set two points within a $2m \times 2m$ square in front of the robot, with each height ranging from 0.1 meters to 2.0 meters. The distance between the points is less than 1 meter. The humanoid needs to touch each point with one hand. 
    \item \texttt{Shelf Touch:} We randomly set a point inside a multi-layer shelf, with height ranging from 0.1 meters to 2.0 meters. The humanoid is asked to touch the point without causing a collision with the shelf. 
    \item \texttt{Box Pickup:} The box is randomly placed within a $2m \times 2m$ square in front of the robot, with height ranging from 0.2 meters to 1.2 meters. The humanoid is asked to lift the box to a height of 1.4 meters.
\end{itemize}

\subsubsection{Experiment Metrics}
We use two metrics:
\begin{itemize}
    \item \textbf{Success Rate}: For point-touch tasks, \textbf{Success Rate} records the percentage of trials that humanoids successfully touch \textit{each} target point within 5 cm. For Box Pickup, \textbf{Success Rate} means the percentage of trials that humanoids successfully lift the box above 1.3 meters.
    \item \textbf{Distance Error}: For point-touch tasks, \textbf{Distance Error} is the averaged closest distance between the humanoid's end effector and the corresponding target point in a touch. For Box Pickup, \textbf{Distance Error} is the closest distance between the box and 1.4 meters height.
\end{itemize}

In simulation, all metrics are averaged over 10000 trials. 
We also measure the sim2real transferability. For sim2real transferable methods, we also report real-world results averaged over 10 trials.

\subsubsection{Baselines}
We choose three baseline methods, including model-free, model-based, and hierarchical RL:
\begin{itemize}
    \item \textbf{Vanilla PPO}~\cite{ppo}: We implement a vanilla PPO without any manually designed prior.
    \item \textbf{Vanilla DreamerV3}~\cite{hafner2023mastering}: We implement a vanilla DreamerV3 without any manually designed prior.
    \item \textbf{HumanoidBench}~\cite{sferrazza2024humanoidbench}: We adopt the hierarchical RL framework in HumanoidBench~\cite{sferrazza2024humanoidbench}, whose low-level policy is a two-hand reaching policy.
\end{itemize}
\subsubsection{Experiment Results}

\begin{figure}
    \centering
    \vspace{0.1cm}
    \includegraphics[width=0.48\textwidth]{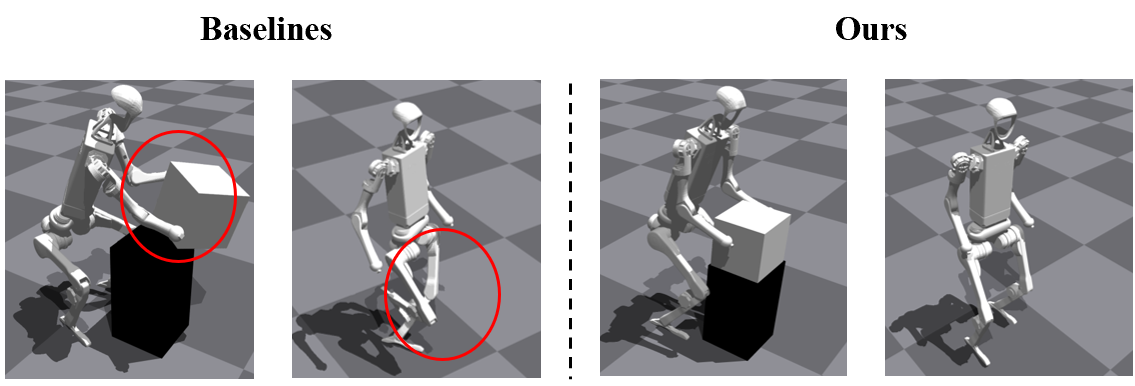}
    \caption{We compare the motor behaviors with baseline methods. Our method significantly helps sim2real transfer.}
    \label{fig:ppo}
\vspace{-6mm}
\end{figure}

The results are shown in Table~\ref{table:unleash}. $\text{R}^{2}\text{S}^{2}$ exceeds baseline methods in task accomplishment performance and sim2real transferability. Vanilla RL methods without any prior, whether model-based or model-free, struggle to learn to accomplish even simple tasks and lack sim2real transferability due to unstable motor behaviors. For the hierarchical method in HumanoidBench~\cite{sferrazza2024humanoidbench}, although it introduces auxiliary rewards that enable successful completion of point-touch tasks, the heavy vibration significantly degrades its performance in other more challenging tasks and prevents its transfer to the real world. The comparison of motor behaviors is shown in Figure~\ref{fig:ppo}.

\subsection{Evaluation of Real-world-Ready Skill Space}
\label{sec:rrss}
In this part, we want to find out how our proposed $\text{R}^{2}\text{S}^{2}$ helps goal-reaching WBC tasks and evaluate the effectiveness of each of our major designs. We again select the four tasks mentioned above and compare the performance of different methods on these tasks.

\subsubsection{Experiment Setting and Metrics}
We reuse the experiment setting and metrics in Section~\ref{sec:unleash}. In this part, we run all experiments in simulation.

\subsubsection{Baselines}
We ablate on different components of our $\text{R}^{2}\text{S}^{2}$ and choose the following baselines:
\begin{itemize}
\item \textbf{$\text{R}^{2}\text{S}^{2}$ w/o PS (Primitive Skills)}: We do not use pre-trained individual primitive skills. Instead, we train a multi-skill controller from scratch to track different commands. We adopt this baseline to verify the necessity of decoupling the multi-skill training to avoid performance degradation.
\item \textbf{$\text{R}^{2}\text{S}^{2}$ w/o SE (Skill Expansion)}: During heterogeneous skill ensembling, we only use IL to train the student policy. In this setting, the student policy is not encouraged to explore new skills with RL. We adopt this baseline to mainly verify the importance of expanding coordination and transition capability.
\item \textbf{$\text{R}^{2}\text{S}^{2}$ w/ Seq-ILRL (Sequential IL and RL)}: We implement a sequential strategy to combine IL and RL in the student policy training process. We first use only IL for pretraining followed by exclusive RL fine-tuning. We adopt this baseline to validate the necessity of maintaining supervision signals throughout the entire training process in prevent of catastrophic forgetting. 
\item \textbf{$\text{R}^{2}\text{S}^{2}$ w/o LS (Latent Space)}: We implement an MLP-based student policy to ensemble skills from multiple teacher policies. In this setting, though the primitive skills are ensembled (i.e., coordination and transition are learned), the high-level planning policy still needs to output skill indicator and the command in the primary mismatched command space. We adopt this baseline to evaluate the effectiveness of our latent skill space. This baseline can also be seen as a comparison with recent works AMO~\cite{li2025amoadaptivemotionoptimization} and HOMIE~\cite{ben2025homie}, which use the primary command space.
\end{itemize}

\subsubsection{Experiment Results}

\begin{figure}
    \centering
    \vspace{0.1cm}
    \includegraphics[width=0.48\textwidth]{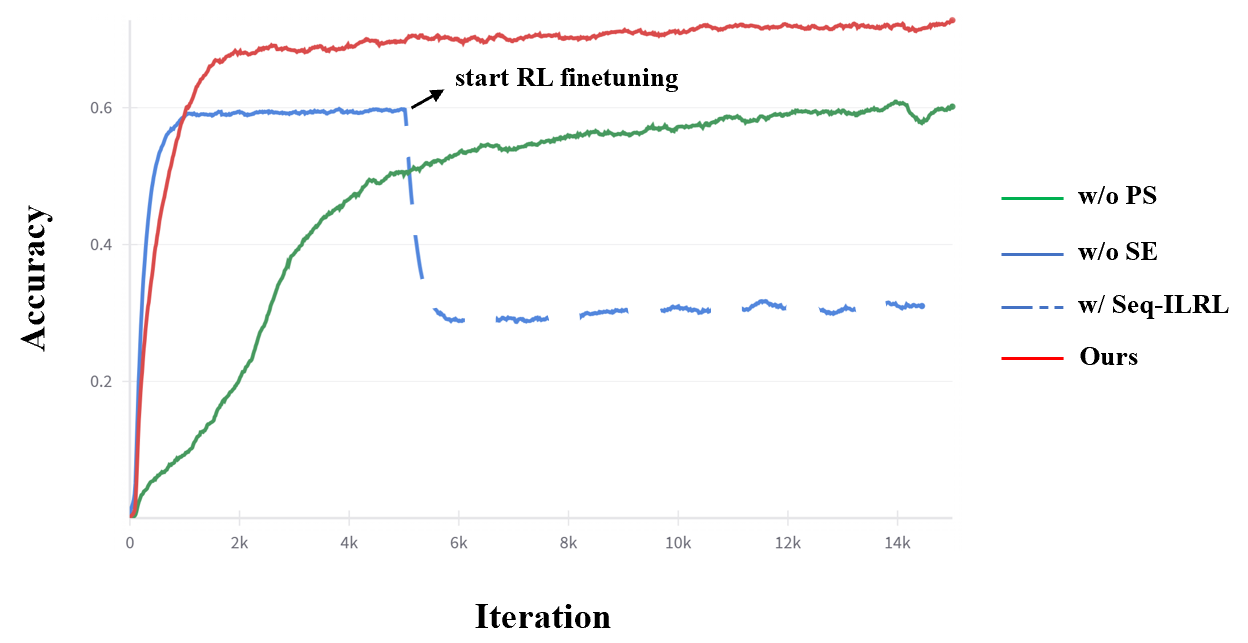}
    \caption{We compare the normalized skill command tracking accuracy.}
    \label{fig:accuracy}
\vspace{-6mm}
\end{figure}

We report the results in Table~\ref{table:rrss}.
For $\text{R}^{2}\text{S}^{2}$ w/o PS, $\text{R}^{2}\text{S}^{2}$ w/o SE, $\text{R}^{2}\text{S}^{2}$ w/ Seq-ILRL, the poor task performance is mainly attributed to insufficient learning of low-level skills, making it difficult for the planning policy to select suitable skills. We visualize the normalized skill command tracking accuracy in Figure~\ref{fig:accuracy}. For $\text{R}^{2}\text{S}^{2}$ w/o PS, the low-level controller cannot learn to track different skill commands very well from scratch. When sim2real transferred, the motor behavior is unstable. For $\text{R}^{2}\text{S}^{2}$ w/o SE, the controller is trained with only IL. It performs unsafely due to the lack of coordination and transition capability. When the skill indicator generated by high-level planners changes and the robot transitions from one lower-body skill to another. It can sometimes fall, thus it is also not sim2real transferable. $\text{R}^{2}\text{S}^{2}$ w/ Seq-ILRL suffers from catastrophic forgetting. Once RL fine-tuning begins, the policy's performance rapidly deteriorates, and it forgets the knowledge acquired during the IL pre-training stage, ultimately performing even worse than training from scratch.

For $\text{
R}^{2}\text{S}^{2}$ w/o LS, the skill command tracking accuracy is not affected. The performance degradation is mainly attributed to the difficulty of high-level planner learning. Sampling from raw combination of skill indicators and mismatched command spaces leads to significantly less efficient RL exploration for high-level planners. Compared with $\text{
R}^{2}\text{S}^{2}$ w/o LS, our latent space provides a more structural action space for multi-skill planning.

\section{Teleoperation system}
As a beneficial side effect, the ensemble student policy $\pi^{\text{ensem}}(a_{t}|s_{t},g_{t})$ can also be utilized to build a humanoid whole-body teleoperation system. Inspired by~\cite{cheng2024open}, we mount a single stereo RGB camera on the robot head. During teleoperation, the VR device worn by the user receives streaming real-time, ego-centric robot observations. The hand pose is captured by the VR device and retargeted to humanoid hand via Inverse Kinematics (IK). We design a pair of split joysticks. The user can hold one joystick in each hand and press the buttons on each joystick to control the lower-body movement of the robot and hand opening/closing without affecting the hand pose. 

\section{conclusion and limitation}
In this work, we propose $\text{R}^{2}\text{S}^{2}$, a structural skill prior to help the execution of autonomous WBC tasks in an efficient and sim2real transferable manner. 
Although we believe our method can be extended to more general and complex whole-body control systems, currently its limitation is obvious: 1) The number of control interfaces and primitive skills is now limited. Incorporating more interfaces (e.g., torso yaw control) and skills is essential for applications in more complex real-world scenarios. 2) Although we achieve seamless coordination and transition at skill ensembling stage, how to blend spatially overlapping skills is still challenging. 3) For now, we rely on the motion capture system to understand the relationship between the robot and the interaction target. Incorporating a visual module is necessary for more general scenarios.

\ifCLASSOPTIONcaptionsoff
  \newpage
\fi

% trigger a \newpage just before the given reference
% number - used to balance the columns on the last page
% adjust value as needed - may need to be readjusted if
% the document is modified later
%\IEEEtriggeratref{8}
% The "triggered" command can be changed if desired:
%\IEEEtriggercmd{\enlargethispage{-5in}}

% references section

% can use a bibliography generated by BibTeX as a .bbl file
% BibTeX documentation can be easily obtained at:
% http://mirror.ctan.org/biblio/bibtex/contrib/doc/
% The IEEEtran BibTeX style support page is at:
% http://www.michaelshell.org/tex/ieeetran/bibtex/
%\bibliographystyle{IEEEtran}
% argument is your BibTeX string definitions and bibliography database(s)
%\bibliography{IEEEabrv,../bib/paper}
%
% <OR> manually copy in the resultant .bbl file
% set second argument of \begin to the number of references
% (used to reserve space for the reference number labels box)
\bibliographystyle{IEEEtran}
\bibliography{IEEEabrv_old}

% biography section
% 
% If you have an EPS/PDF photo (graphicx package needed) extra braces are
% needed around the contents of the optional argument to biography to prevent
% the LaTeX parser from getting confused when it sees the complicated
% \includegraphics command within an optional argument. (You could create
% your own custom macro containing the \includegraphics command to make things
% simpler here.)
%\begin{IEEEbiography}[{\includegraphics[width=1in,height=1.25in,clip,keepaspectratio]{mshell}}]{Michael Shell}
% or if you just want to reserve a space for a photo:

% insert where needed to balance the two columns on the last page with
% biographies
%\newpage

% You can push biographies down or up by placing
% a \vfill before or after them. The appropriate
% use of \vfill depends on what kind of text is
% on the last page and whether or not the columns
% are being equalized.

%\vfill

% Can be used to pull up biographies so that the bottom of the last one
% is flush with the other column.
%\enlargethispage{-5in}

% that's all folks
\end{document}